\documentclass[10pt,twocolumn,letterpaper]{article}

\usepackage{iccv}              % To produce the CAMERA-READY version
\usepackage{graphicx,verbatim}
\usepackage[table,xcdraw]{xcolor}
\usepackage{booktabs}
\usepackage{amssymb}
\usepackage{multirow}
\definecolor{iccvblue}{rgb}{0.21,0.49,0.74}
\usepackage[pagebackref,breaklinks,colorlinks,allcolors=iccvblue]{hyperref}
\def\paperID{23} % *** Enter the Paper ID here
\def\confName{ICCV}
\def\confYear{2025}

\title{CECT-Mamba: a Hierarchical Contrast-enhanced-aware Model for Pancreatic Tumor Subtyping from Multi-phase CECT}

\author{
    Zhifang Gong$^{1}$, 
    Shuo Gao$^{1}$,
    Ben Zhao$^{1,2}$,  
    Yingjing Xu$^{3}$,
    Yijun Yang$^{4}$,
    Shenghong Ju$^{1,2}$,
    Guangquan Zhou$^{1}$\footnotemark[2]\\
    $^1$Southeast University~~~
    $^2$Zhongda Hospital, Medical School of Southeast University~~~\\
    $^3$Zhejiang University~~~
    $^4$The Hong Kong University of Science and Technology (Guangzhou)~~~\\
}
\begin{document}
\maketitle
\begin{abstract}
Contrast-enhanced computed tomography (CECT) is the primary imaging technique that provides valuable spatial-temporal information about lesions, enabling the accurate diagnosis and subclassification of pancreatic tumors. However, the high heterogeneity and variability of pancreatic tumors still pose substantial challenges for precise subtyping diagnosis. Previous methods fail to effectively explore the contextual information across multiple CECT phases commonly used in radiologists' diagnostic workflows, thereby limiting their performance. In this paper, we introduce, for the first time, an automatic way to combine the multi-phase CECT data to discriminate between pancreatic tumor subtypes, among which the key is using Mamba with promising learnability and simplicity to encourage both temporal and spatial modeling from multi-phase CECT. Specifically, we propose a dual hierarchical contrast-enhanced-aware Mamba module incorporating two novel spatial and temporal sampling sequences to explore intra and inter-phase contrast variations of lesions. A similarity-guided refinement module is also imposed into the temporal scanning modeling to emphasize the learning on local tumor regions with more obvious temporal variations. Moreover, we design the space complementary integrator and multi-granularity fusion module to encode and aggregate the semantics across different scales, achieving more efficient learning for subtyping pancreatic tumors. The experimental results on an in-house dataset of 270 clinical cases achieve an accuracy of 97.4\% and an AUC of 98.6\% in distinguishing between pancreatic ductal adenocarcinoma (PDAC) and pancreatic neuroendocrine tumors (PNETs), demonstrating its potential as a more accurate and efficient tool.
\end{abstract}    
\section{Introduction}
\label{sec:intro}

\begin{figure*}[ht]
  \includegraphics[width=\textwidth]{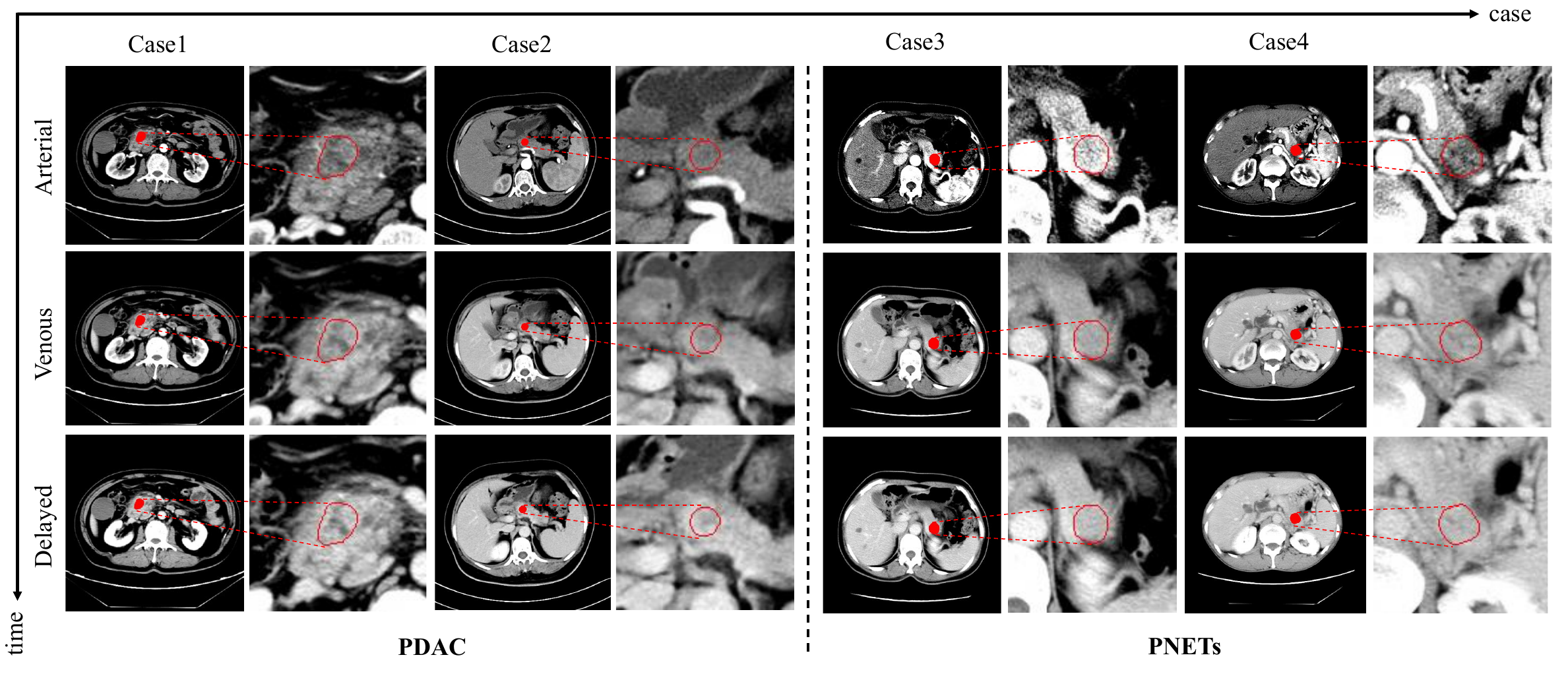}
  \caption{Example of PDAC and PNETs. Columns 1–4 show two cases of PDAC, and Columns 5–8 show two cases of PNETs, with each case occupying two columns: the left column displays the full original image, and the right column shows an enlarged view of the tumor region. Each row corresponds to a different phase—arterial, venous, and delayed phases, respectively. Notably, PDAC and PNETs exhibit certain inter-class similarities in both appearance and texture on CECT images, while also presenting both typical and atypical enhancement patterns across phases.} 
  \label{fig_intro}
\end{figure*}

Pancreatic tumors, typically diagnosed at advanced stages, result in poor prognosis and limited treatment options, posing a significant threat to human health~\cite{num2}. Contrast-enhanced computed tomography (CECT) is the primary imaging modality for accurately diagnosing and subclassifying pancreatic tumors~\cite{num3}, corresponding to different clinical treatment and prognosis strategies. For instance, pancreatic neuroendocrine tumors (PNETs) tend to respond more favorably to surgical intervention and targeted chemotherapy, while pancreatic ductal adenocarcinoma (PDAC) typically exhibits a more aggressive progression and limited responsiveness to conventional treatments~\cite{garrido2015pancreatic,num2}. However, as shown in Fig.~\ref{fig_intro}, the high inter-class visual and texture similarity, along with the presence of atypical enhancement patterns in certain cases, poses substantial challenges for accurate tumor subtyping based on CECT~\cite{num4}. Radiologists often rely on valuable spatial-temporal information about lesions offered by the enhancement patterns across different phases of CECT to determine pancreatic tumor subtypes~\cite{num3}, which is heavily time-consuming and subjective. Therefore, an automatic pancreatic tumor subclassification method is desirable to guide accurate clinical treatment strategies and improve diagnostic efficiency for radiologists.

Existing studies have progressed in automatically diagnosing tumors imperceptible to the human eye~\cite{num9,num10,num11,yang2023diffmic,yang2025diffmic,yang2023mammodg,yang2021hcdg}. Nonetheless,  these methods fail to explore the enhancement variation across CECT phases, restring the application in pancreatic tumor classification. Recently, some research has attempted to integrate information from multiple phases to capture salient lesion characteristics at various time points simultaneously~\cite {gao2021deep,szeskin2023liver,num15,num16,yang2025medical,yang2024genuine}.~\cite{num15} separately extracted features from three-phase CECT, followed by the concatenation into ConvLSTM~\cite{num13} to predict pancreatic tumor survival. Similarly,~\cite{num16,ali2025objective} employed an ordered recursive convolutional concatenation operation to extract temporal features across multi-phase CT images for liver tumor classification. The performance of these methods is still limited as their computation becomes complexity quadratic with the 3D volume size, thus restricting the ability to model the global temporal contextual information across different CECT phases.

Mamba~\cite{num17}, derived from state-space models (SSMs)~\cite{num18}, shows great potential for its efficiency in analyzing long sequences of data, which is recognized to improve the ability of long-range dependency modeling. Vision Mamba~\cite{num19} adopts bidirectional Mamba blocks to capture global contextual information for each image patch from image sequences, overcoming the computational limitations in high-dimensional data. SegMamba~\cite{num20} integrates Mamba into a U-shaped encoder-decoder architecture for 3D data segmentation, exhibiting the ability in high-efficiency modeling whole-volume features. Similarly,~\cite{yang2024vivim,yang2025vivim} designs a temporal Mamba block for long-term temporal encoding to realize efficient medical video segmentation. Despite these advances, the previous Mamba-based scanning sequences remain limited in their ability to achieve more accurate and scalable modeling of dynamic medical data, thus requiring further optimizations for effectively capturing complex temporal dependencies arising from dynamic change patterns across different scales.

Motivated by this, this study presents an automatic classification way based on Mamba, dubbed  \textbf{CECT-Mamba}, to discriminate pancreatic tumor subtypes using multi-phase CECT data. Our framework has three highlights: 

1) To the best of our knowledge, \textit{it is the first work to explore Mamba for 3D multi-phase data analysis.} We construct a multi-phase CECT dataset with arterial, portal venous, and delayed phases for pancreatic tumor subclassification. 

2) We propose a \textbf{Dual Hierarchical Contrast-enhanced-aware Mamba (DHCM)} module incorporating two novel complementary sampling sequences to analyze intra-phase and inter-phase spatial-temporal variations. Specifically, the spatial scanning sequence is designed to examine intra-phase spatial dependencies in conjunction with global temporal changes. In contrast, temporal scanning focuses on exploring inter-phase temporal dependencies of local regions. Additionally, we introduce a \textbf{Similarity-guided Refinement (SimR)} module within the temporal scanning sequence to elevate the scrutinization of localized tumor regions that exhibit highly variable across phases. 

3) The \textbf{Space Complementary Integrator (SCI)} and \textbf{Multi-Granularity Fusion (MGF)} modules are also employed to compress and aggregate the semantic information of spatial-temporal features across multiple scales, respectively, for more efficient learning of 3D data. Experiment results showed that our model can process rich spatial-temporal information from multi-phase data and identify tumor subtypes accurately and efficiently.

%-------------------------------------------------------------------------

%-------------------------------------------------------------------------

\section{Related work}
\label{sec:relatedwork}

\subsection{Pancreatic Tumor Diagnosis}
Recent advances in computer-aided diagnosis have led to various approaches for automated pancreatic tumor recognition and classification. ~\cite{sujatha2021soft} proposed a method that utilizes statistical texture features for pancreatic tumor detection, followed by deep wavelet neural networks for tumor classification.~\cite{xuan2020detection} introduced a hierarchical convolutional neural network to improve the accuracy of pancreatic tumor diagnosis. ~\cite{asadpour2021pancreatic} presented a cascaded architecture that combines an elastic atlas with a multi-resolution CNN for tumor extraction in patients with adenocarcinoma.~\cite{gao2020performance} explored the use of generative adversarial networks (GANs) to distinguish pancreatic tumors on contrast-enhanced MRI.~\cite{vaiyapuri2022intelligent} integrated Gabor filtering for preprocessing, multi-level threshold segmentation optimized by the Emperor Penguin Optimizer, MobileNet for feature extraction, and an autoencoder for classification, further enhanced by the Multi-Leader Optimizer to improve classification accuracy.
However, despite these advances, their diagnostic performance remains limited. To the best of our knowledge, no prior work has explored an automated method that leverages multi-phase contrast-enhanced CT images for subtype classification of pancreatic tumors.

\subsection{Medical Image Analysis based on CECT}
Several deep learning methods use multi-phase CT scans for the image analysis of anatomical structures that yield better results than single-input methods~\cite{xu2021pa, ouhmich2019liver}.~\cite{xu2021pa} introduces a phase attention mechanism, leveraging the arterial phase to facilitate the segmentation of the portal venous phase. This method yields an improvement of 10.4\% and 21.5\% with respect to standalone detection in either scan. To leverage the temporal characteristics of CECT across different phases, ~\cite{gao2021deep} leverages CECT by combining a CNN to extract spatial features from each phase and a gated RNN to capture temporal enhancement patterns across phases, achieving accurate differentiation of malignant hepatic tumors.~\cite{num15} extracted features from three-phase CECT images using separate CNNs and concatenated them as input to a ConvLSTM to predict pancreatic tumor survival. Similarly,~\cite{num16} introduced an ordered recursive convolutional concatenation operation to capture temporal features across multi-phase CT images for liver tumor classification. However, these methods lack the ability to capture global context and fine-grained enhancement variations at multiple scales, which are critical for accurate subtyping of pancreatic tumors.

\subsection{Mamba}
Recently, State Space Models (SSMs)~\cite{num18} have demonstrated remarkable efficiency in managing long-range dependencies within language sequences by leveraging state space transformations~\cite{gu2021combining}. S4~\cite {gu2021efficiently} introduced a structured state space sequence model, which exploits the advantages of linear complexity to capture long-range dependencies. Mamba~\cite{num17}, derived from SSMs, integrates efficient hardware design and a selection mechanism employing parallel scanning (S6), thereby surpassing Transformers in processing a wide range of natural language sequences. It shows great potential for its efficiency in analyzing long sequences of data, which is recognized to improve the ability of long-range dependency modeling.

Mamba is not only widely used in natural language processing but also excels in capturing long-range dependencies in image processing. 
% Vision Mamba~\cite{num19} adopts bidirectional Mamba blocks to capture global contextual information for each image patch from image sequences, overcoming the computational limitations in high-dimensional data. 
Vision Mamba~\cite{num19} and VMamba~\cite{liu2024vmamba} employ multi-directional Mamba blocks, extending beyond vanilla Mamba’s single-scan approach, to effectively capture long-range dependencies for each image patch from image sequences.
% SegMamba~\cite{num20} integrates Mamba into a U-shaped encoder-decoder architecture for 3D data segmentation, exhibiting the ability in high-efficiency modeling whole-volume features. 
U-Mamba~\cite{ma2024u}, SegMamba~\cite{num20} and SwinUMamba~\cite{liu2024swin} have effectively incorporated Mamba blocks as plug-in components within convolutional neural network-based architectures, achieving strong performance on biomedical segmentation benchmarks.
Similarly, Vivim and other variants~\cite{yang2025vivim,xu2024lgrnet} design a temporal Mamba block for long-term temporal encoding to realize efficient medical video segmentation. RainMamba~\cite{wu2024rainmamba} applies a novel Hilbert scanning mechanism to better capture sequence-level local information on videos. 
% Despite these advances, the previous Mamba-based scanning sequences remain limited in their ability to achieve more accurate and scalable modeling of dynamic medical data, thus requiring further optimizations for effectively capturing complex temporal dependencies arising from dynamic change patterns across different scales.
While promising, current Mamba-based methods struggle with fine-grained temporal modeling and scalability in dynamic medical imaging. Enhancing their capacity to learn multi-scale spatiotemporal dependencies remains critical for handling real-world clinical variability.
\section{Method}
\label{sec:formatting}

\begin{figure*}[ht]
  \includegraphics[width=\textwidth]{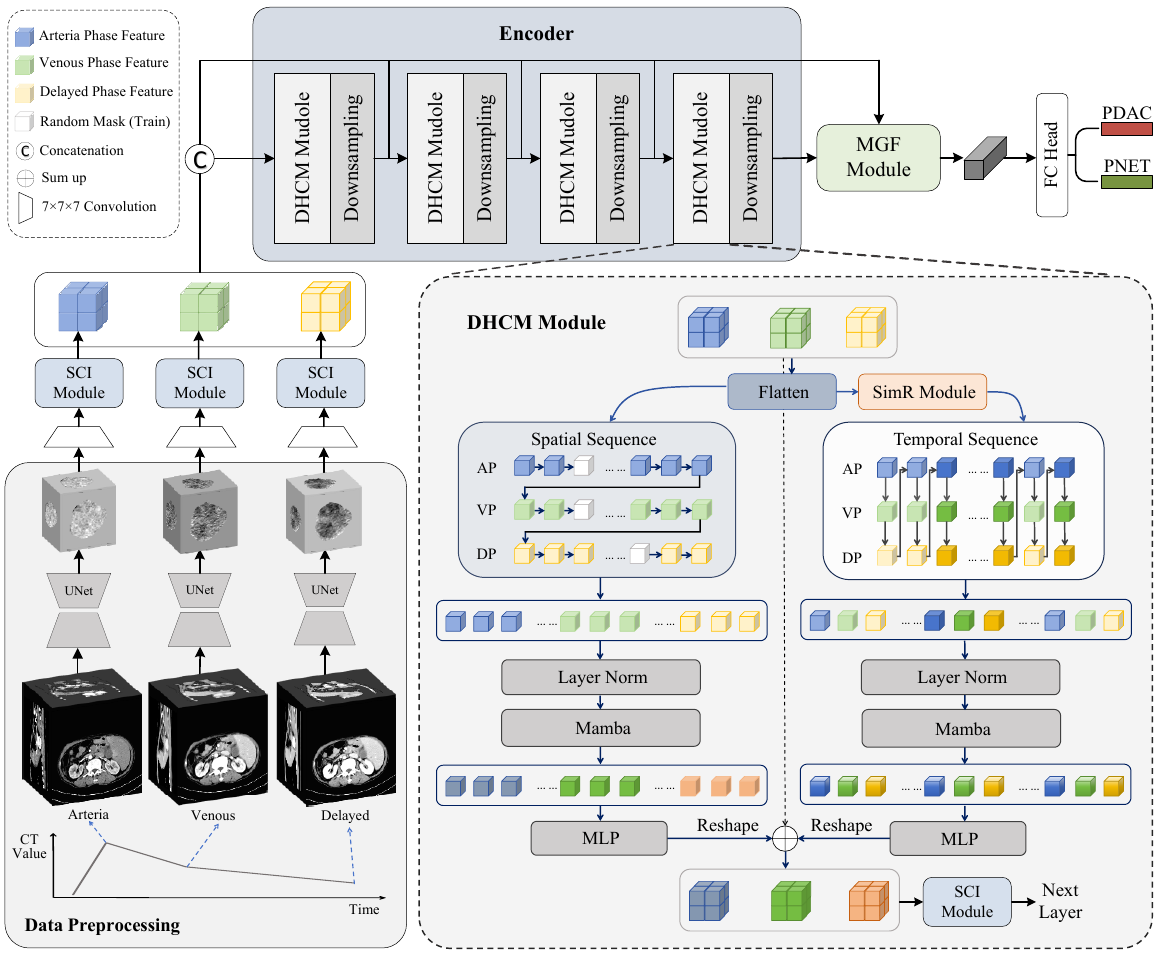}
  \caption{\textbf{The overview of CECT-Mamba.} The multi-phase CT images are first preprocessed to capture the tumor region. Then, the 3D features of each phase are extracted using the CNN and the SCI module. These features are then aggregated into the DHCM modules and downsampling blocks to model the temporal-spatial dependency. Finally, the MGF module fuses the multi-scale features to determine the tumor subtype.} 
  \label{fig1}
\end{figure*}

CECT-Mamba, depicted in Fig.~\ref{fig1}, is a multi-stage framework with the Mamba block as its core. Given the CECT 3D data of the arterial, venous, and delayed phases, the framework begins with an encoder-decoder model to localize and crop pancreatic tumor regions from the whole volume for each phase, followed by a depth-wise large kernel size convolution and the SCI (Sect. \ref{sec:SCI}) module to encode the 3D feature in conjunction with spatially complementary details enhancement using different receptive fields. Subsequently, the four-layer encoder consisting of DHCM (Sect. \ref{sec:DHCM}) modules and down-sampling blocks processes all latent features from the three phases with two novel spatial and temporal sampling sequences to explore intra and inter-phase contrast variations of lesions at different granularities. Finally, all obtained hierarchical temporal-spatial features are fed into the MGF (Sect. \ref{sec:MGF}) module to aggregate tumor salient features across various scales for classifying the pancreatic tumor subtype.
%-------------------------------------------------------------------------
% \begin{figure}[ht]
% \includegraphics[width=\textwidth]{FIGURE/method_1_0229.pdf}
% \caption{\textbf{The overview of CECT-Mamba.} The multi-phase CT images are first preprocessed to capture the tumor region. Then, the 3D features of each phase are extracted using the CNN and the SCI module. These features are then aggregated into the DHCM modules and downsampling blocks to model the temporal-spatial dependency. Finally, the MGF module fuses the multi-scale features to determine the tumor subtype.} \label{fig1}
% % \vspace{-0.3cm}
% \end{figure}

\subsection{Dual Hierarchical Contrast-enhanced-aware Mamba (DHCM)}
\label{sec:DHCM}
To effectively capture tumors' spatiotemporal contrast enhancement features at different hierarchies of granularity, we design the DHCM module, which leverages the inherent position-sensitive characteristics of Mamba to compute hierarchical dependencies for the contrast-enhanced variations of the global space and local regions. As shown in Fig.~\ref{fig1}, We first flatten the 3D features of each phase, $F=\{f^\mathrm{A},f^\mathrm{V},f^\mathrm{D}\}$, into 1D sequences $S=\{s^\mathrm{A},s^\mathrm{V},s^\mathrm{D}\}$, where $A$, $V$ and $D$ denote the arterial, venous, and delayed phases, respectively. Then, we construct two complementary embedding sequences by leveraging intra- and inter-phase concatenation orders of tokens in $\{s^\mathrm{A},s^\mathrm{V},s^\mathrm{D}\}$ to enable simultaneous spatial and temporal modeling.
% \vspace{-0.4cm}

\begin{figure*}[t]
\includegraphics[width=\textwidth]{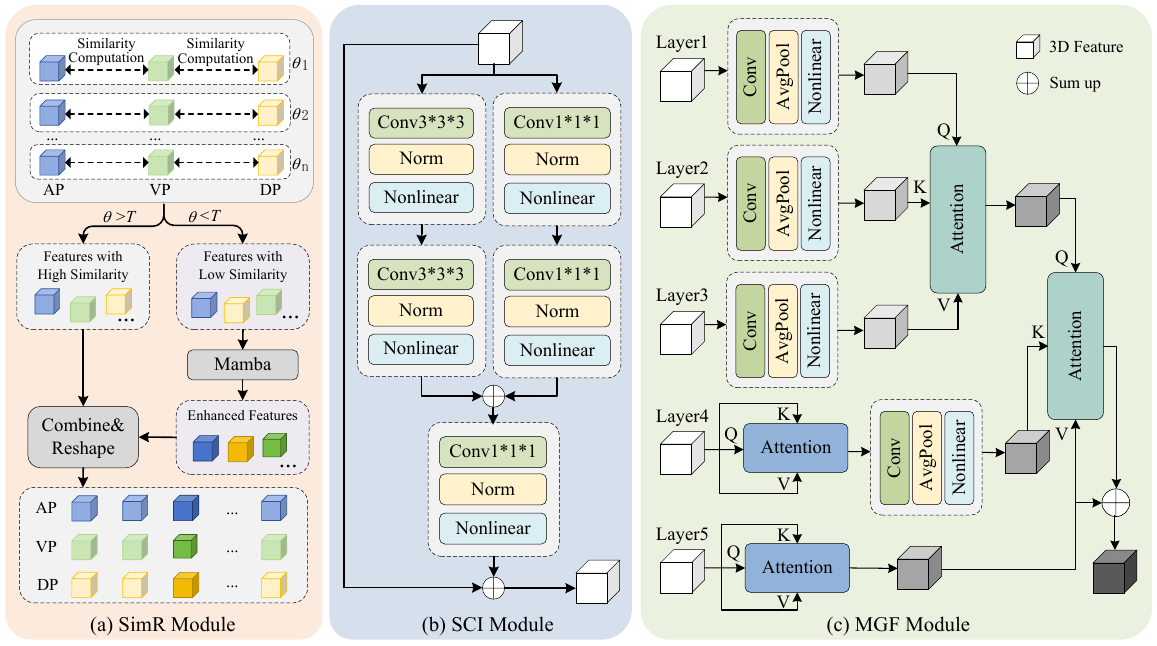}
\caption{(a) The Similarity-guided Refinement module enhances embeddings with low similarity across CT phases. (b) The Space Complementary Integrator module enriches space information of 3D features. (c) The Multi-Granularity  Fusion module fuses features from different scales.} \label{fig2_method}
% \vspace{-0.3cm}
\end{figure*}

\subsubsection{Spatial Modeling}
% \textbf{Spatial modeling.} 
The flattened tokens are first concatenated within each phase to construct the tumor's whole-volume information and then sequentially combined across phases. This embedding scanning sequence is subsequently fed into the Mamba layer for modeling the global variation pattern of the whole tumor region. In addition, during the training phase, we apply random masking to the embedding sequences to enhance the robustness of the model. The computation process is defined as:
\begin{equation}
    \mathrm{x_{s}}^{\prime}=Mamba\left(LN\left(\mathcal{M}(\mathrm{x_{s}})\right)\right),\mathrm{x_{s}}^{\prime\prime}=MLP(\mathrm{x_{s}}^{\prime}),
\end{equation}
where $\mathrm{x_{s}}$ represents the spatial embedding scanning sequence, $\mathcal{M}$ represents the random masking, $LN$ denotes layer normalization, $Mamba$ refers to the Mamba layer, which captures global features, and $MLP$ denotes multi-layer perceptron.

\subsubsection{Temporal Modeling}
% \textbf{Temporal modeling.} 
Temporal modeling aims to capture the fine-grained enhancement dynamics of tumors across different phases at the voxel level. Before constructing the temporal scanning order, a Similarity-guided Refinement (SimR) module is applied to select and enhance localized embeddings that exhibit significant variation across phases, as illustrated in Fig.~\ref{fig2_method}(a). The similarity calculation can be defined as:
\begin{equation}
    \theta_{\mathrm{i}}=\delta(F_{i}^{A},F_{i}^{V})+\delta(F_{i}^{V},F_{i}^{D}),
\end{equation}
where $\delta$ denotes cosine similarity computation. $F_{i}^{A}$, $F_{i}^{V}$, $F_{i}^{D}$ denotes the 
$i$-th embedding of the sequence from the arterial, venous, and delayed phases, respectively. $\theta_{\mathrm{i}}$ denotes the similarity score of the $i$-th embedding.

According to $\theta_{\mathrm{i}}$, embeddings are ranked into high-similarity and low-similarity subsets, with the low similarity fed into the Mamba layer for enhancement. 
% After reconstructing the original positions of the enhanced and high-similarity embeddings, a temporal scanning sequence is formed by linking embeddings across the three phases. 我们将同一个位置的token在三期之间先做顺序的串联,再按照空间信息依次连接。This sequence is then fed into the Mamba layer to perform feature interaction for the same localized tumor region, capturing 肿瘤区域在多层次细粒度下的continuous inter-phase variations.
After reconstructing the original positions of the enhanced low-similarity embeddings, we form a temporal scanning sequence by first linking the tokens at the same location in temporal order, and then concatenating them spatially. This sequence is subsequently fed into the Mamba layer to enable feature interaction within the same localized tumor region, allowing the model to capture continuous inter-phase enhancement variations of the tumor at multiple levels of fine granularity. The calculation can be defined as:
\begin{equation}
\mathrm{x_{t}}^{\prime}=Mamba\left(LN\left((\mathrm{x_{t}})\right)\right),\mathrm{x_{t}}^{\prime\prime}=MLP(\mathrm{x_{t}}^{\prime}),
\end{equation}
where $\mathrm{x_{t}}$ represents the temporal embedding scanning sequence.

Finally, the feature sequences from spatial and temporal modeling are restored to their original 3D shapes, summed together, and then passed through the down-sampling block for the subsequent processing.

\subsection{Space Complementary Integrator (SCI)}
\label{sec:SCI}
To mitigate the significant loss of spatial information after passing through the Mamba layer, we introduce the SCI module.  This module enriches the complementary details across different feature levels to improve the model’s representational capacity. As illustrated in Fig.~\ref{fig2_method}(b), 3D features are processed through two parallel paths: One path employs two $3$$\times$$3$$\times$$3$ convolution blocks to encode channel-wise spatial context. Each block consists of a convolution, a normalization layer, and a nonlinear activation layer. Another path applies two $1$$\times$$1$$\times$$1$ convolution blocks to aggregate pixel-wise cross-channel context. Finally, the enhanced spatial features are obtained through residual connections. The computation process is defined as:
\begin{equation}
    SCI(z)=z+C^{1}(C^{3}\cdot C^{3}(z)+C^{1}\cdot C^{1}(z)),
\end{equation}
where z denotes the input 3D features and C denotes the 3D convolution block.

\subsection{Multi-Granularity Fusion (MGF)}
\label{sec:MGF}
% The HAF module effectively facilitates the interaction between fine details and global context, thereby improving the global and local perception of the model in CECT images. 
To integrate the spatial-temporal semantics across multiple scales, the MGF module fuses features by transferring detail-rich information from shallow layers to deeper abstract features. As illustrated in Fig.~\ref{fig2_method}(c), we consider five hierarchical feature levels $F_{Li},i=\{1,2,3,4,5\}$ from shallow to deep. For the shallow features from the first three layers, the MGF module first applies convolutional blocks, consisting of convolution, pooling, and nonlinear activation layers, to align the input features to a unified dimensionality $F_{Li}^{\prime}$, followed by cross-scale multi-head self-attention~\cite{num21}, producing the fused representation $F{^{\prime{\prime}}}$. For deeper features $F_{L4}, F_{L5}$, intra-scale self-attention is applied to filter out noise and redundancy introduced by multiple downsamplings, followed by cross-scale attention with $F{^{\prime{\prime}}}$ to generate the final fused feature.
The computation process of the MLF module can be defined as:
\begin{equation}
    F^{\prime{\prime}} = CA(F_{L1}^{\prime},F_{L2}^{\prime},F_{L3}^{\prime}),
\end{equation}
\begin{equation}
    MGF(F)=CA(F^{\prime{\prime}},SA(F_{L4})^{\prime},SA(F_{L5})),
\end{equation}
where $F$ represents the multi-scale fused feature, $CA$ denotes the cross-attention mechanism, and $SA$ refers to the multi-head self-attention mechanism.

\section{Experiments}

\subsection{Dataset}
\textbf{Dataset Construct.} We establish a CECT dataset comprising 270 patients, including 184 patients with PDAC and 96 patients with PNETs. Each patient underwent dynamic CECT scans of three phases (arterial, venous, and delayed), forming a total of 810 3D images. All images from the other phases were registered to the arterial phase using Elastix~\cite{num22}. Pancreatic tumors were annotated and calibrated by two abdominal radiologists with three and five years of clinical experience respectively, and the tumor types were confirmed through pathological examination after surgical resection. The CT scans were acquired from January 2015 to December 2020, with all sensitive patient information removed.

\textbf{Dataset Analysis.} All CT scans have a consistent in-plane dimension of 512 $\times$ 512, with the z-axis dimension ranging from 41 to 89 slices and a median of 57. The in-plane spacing varies from 0.609 $\times$ 0.609 mm to 0.977 $\times$ 0.977 mm, with a median of 0.744 $\times$ 0.744 mm, while the z-axis spacing is consistently 5.0 mm.

\subsection{Implementation Details}
For localizing pancreatic tumors, we utilize a pre-trained nnUNet~\cite{num23} configured for high-resolution image processing as the localization network. Once localized, the ROIs are cropped to dimensions of 128 $\times$ 128 $\times$ 32, followed by preprocessing steps including resampling and intensity normalization. Throughout the training process, data augmentation techniques including random flipping and the addition of random noise are applied. In the DHCM module, 50\% of the embeddings within the sequence are randomly masked. In the SimR module, all tokens are ranked based on similarity and then split evenly into high-similarity and low-similarity groups. For the downsampling block, we adopt the downsampling module from UNETR~\cite{hatamizadeh2022unetr}. The model is trained using the Adam optimizer with an initial learning rate of 1e-5. Training spans 100 epochs, with the learning rate progressively reduced using a cosine annealing scheduler to a final value of 1e-7. The entire framework is implemented in PyTorch and executed on NVIDIA 4090 GPUs.

\begin{table*}[t]
\centering
\caption{Comparisons with several state-of-the-art methods with 5-fold cross-validation.  Accuracy (ACC), Area Under the ROC Curve (AUC), F1-Score, Recall, Precision and Inference Time (IT) are reported as our evaluation metrics. The best scores are highlighted in bold.}

\tabcolsep=0.9mm
\begin{tabular}{@{}cccccccc@{}}
\toprule
Method         & Venue                   & ACC (\%)$\uparrow$              & AUC (\%)$\uparrow$              & Recall (\%)$\uparrow$ & Precision (\%)$\uparrow$ & F1-Score (\%)$\uparrow$          & IT$\downarrow$             \\ \midrule
3D-Resnet50~\cite{num25}        & CVPR18     & 87.26 ± 6.13          & 79.96 ± 11.93         & 65.74 ± 20.35         & \textbf{98.67 ± 2.67}           & 76.81 ± 15.08         & \textbf{0.25} \\ \midrule
E3D-LSTM~\cite{num12}        & ICLR18        & 94.19 ± 3.99          & 95.54 ± 5.23          & 88.16 ± 9.80          & 96.16 ± 5.26            & 91.76 ± 6.86          & 8.41          \\
MPBD-LSTM~\cite{num26}      & MICCAI23       & 95.30 ± 3.10          & 95.64 ± 4.71          & 88.73 ± 9.35          & 97.80 ± 4.90            & 93.42 ± 5.63          & 10.70          \\ \midrule
ViT~\cite{num27}        & ICLR20             & 89.37 ± 5.16          & 86.11 ± 7.90          & 72.87 ± 15.95         & 95.39 ± 3.87           & 81.67 ± 10.73         & 1.26          \\
SwinT~\cite{num28}        & ICCV21 & 93.81 ± 3.39          & 94.83 ± 4.68          & 85.81 ± 10.20          & 97.39 ± 3.22           & 90.90 ± 6.28          & 1.07          \\
Uniformer~\cite{num30}       & TPAMI23       & 92.96 ± 3.78          & 93.47 ± 3.24          & 82.28 ± 8.22          & 94.46 ± 5.31            & 87.86 ± 6.63          & 0.97          \\ \midrule
Diff3Dformer~\cite{num29}      & MICCAI24    & 96.78 ± 2.16          & 97.25 ± 1.84          &  90.73 ± 8.07          & 98.47 ± 2.48            & 94.06 ± 5.34          & 0.89          \\ \midrule
\rowcolor[HTML]{EFEFEF}
\textbf{CECT-Mamba} & \textbf{Ours}      & \textbf{97.41 ± 2.77} & \textbf{98.60 ± 2.18} & \textbf{93.13 ± 6.88} & 98.06 ± 2.02  & \textbf{95.48 ± 4.97} & 0.85          \\ \bottomrule
\label{table1}
\end{tabular}
\end{table*}

\begin{figure*}[htp]
\includegraphics[width=\textwidth]{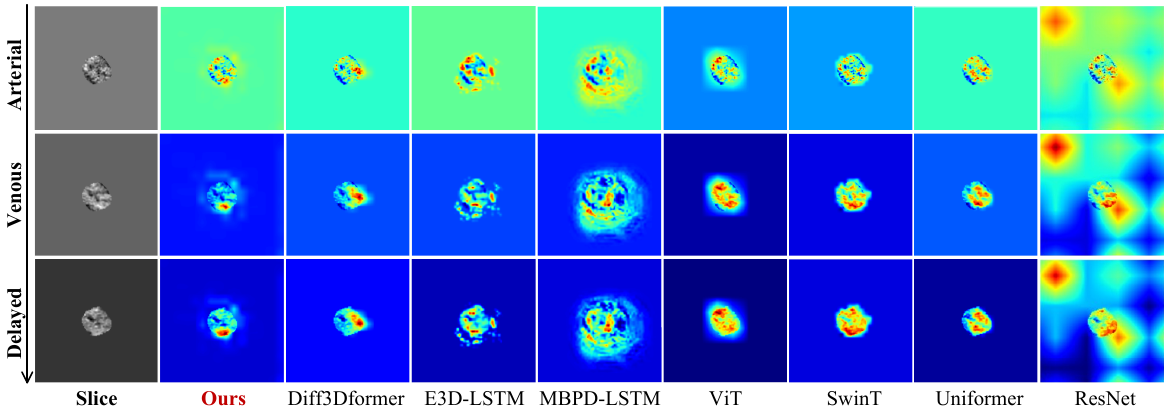}\label{fig3}
\caption{The Grad-CAM visualization of our method with other comparison methods. Lines 1 to 3 represent the tumor region of a patient's CT slice across three phases: arterial, venous, and delayed, respectively. This case represents a challenging example in which the true label, determined by the postoperative pathological gold standard, is a PNET, while radiologists misdiagnosed it as PADC based on CECT images. Among all comparison methods, only our model correctly predicted the subtype and focused more precisely on the tumor’s peripheral region.}
% \vspace{-0.3cm}
\label{fig3}
\end{figure*}

\subsection{Comparison with Other Methods}
\subsubsection{Baselines and Metrics}
% \textbf{Baselines and metrics.} 
We compare our proposed method with seven state-of-the-art methods, including CNN-based method (ResNet3D-50~\cite{num25}), LSTM-CNN hybrid architectures (E3D-LSTM~\cite{num12}, MPBD-LSTM~\cite{num26}), transformer-based methods fine-tuned to accommodate 3D embedding inputs (Vision Transformer~\cite{num27}, Swin Transformer~\cite{num28}), transformer-CNN hybrid architecture (Uniformer~\cite{num30}) and 2.5D transformer-based methods (Diff3Dformer~\cite{num29}). 
% It is also worth noting that Diff3Dformer relies on a pretrained diffusion autoencoder as a feature extractor. This reconstruction-based training process incurs significant computational cost, making it much more resource-intensive than our approach.
For the non-LSTM-based method, the features of the three-phase CT images are fused through concat~\cite{num31} and fed into the classification head. For a fair comparison, we use the same preprocessing to obtain the tumor region. 
% all methods were evaluated using 5-fold cross-validation. The Accuracy (ACC), Area Under the ROC Curve (AUC) and F1-Score metrics are adopted for quantitative comparison on our dataset. We also evaluate inference speed by measuring the average inference time (IT) per 10 cases in seconds.

\begin{table*}[t]
\centering
\caption{Ablation study for different modules with 5-fold cross-validation. In DHCM, $M_S$ denotes the spatial modeling and $M_T$ denotes the temporal modeling. The results indicate a trend of performance improvement as each proposed component is progressively incorporated.}
\tabcolsep=0.9mm
\begin{tabular}{@{}ccccccccccc@{}}
\toprule
\multirow{2}{*}{Methods} & \multicolumn{1}{c}{\multirow{2}{*}{SCI}} & \multicolumn{3}{c}{DHCM}                                                   & \multirow{2}{*}{MGF} & \multirow{2}{*}{ACC (\%)} & \multirow{2}{*}{AUC (\%)} & \multirow{2}{*}{Recall (\%)} & \multirow{2}{*}{Precision (\%)} & \multirow{2}{*}{F1-Score (\%)} \\ \cmidrule(lr){3-5}
                         & \multicolumn{1}{c}{}                     & \multicolumn{1}{c}{$M_S$} & \multicolumn{1}{c}{$M_T$} & \multicolumn{1}{c}{SimR} &                      &                          &                           &                              &                                 &                            \\ \midrule
Basic                    &                                          &                        &                        &                         &                      & 86.67 ± 4.60             & 83.47 ± 7.33              & 58.58 ± 18.25              & 94.82 ± 4.90                & 72.04 ± 12.14             \\
C1                       & $\checkmark$                             &                        &                        &                         &                      & 93.86 ± 5.16             & 92.37 ± 6.06              & 78.30 ± 9.54              & 96.24 ± 3.81	                & 87.28 ± 7.46              \\
C2                       & $\checkmark$                             & $\checkmark$           &                        &                         &                      & 94.82 ± 5.56             & 95.16 ± 3.74              & 87.43 ± 9.22              & 96.04 ± 4.10                & 91.43 ± 6.31              \\
C3                       & $\checkmark$                             & $\checkmark$           & $\checkmark$           &                         &                      & 95.96 ± 5.11             & 96.78 ± 2.08              & 90.60 ± 8.35              & 96.28 ± 2.17                & 93.72 ± 5.82              \\
C4                       & $\checkmark$                             & $\checkmark$           & $\checkmark$           & $\checkmark$           &                      & 96.30 ± 3.51             & 97.22 ± 2.72              & 92.18 ± 7.46              &  96.50 ± 2.12                & 93.31 ± 6.06              \\ \midrule
\rowcolor[HTML]{EFEFEF}
\textbf{Ours}            & $\checkmark$                             & $\checkmark$           & $\checkmark$           & $\checkmark$           & $\checkmark$         & \textbf{97.41 ± 2.77}    & \textbf{98.60 ± 2.18}     & \textbf{93.13 ± 6.88}     & \textbf{98.06 ± 2.02}       & \textbf{95.48 ± 4.97}     \\ \bottomrule
\label{table2}
\end{tabular}
\end{table*}

All methods are evaluated using 5-fold cross-validation. For quantitative comparison on our dataset, we report Accuracy (ACC), Area Under the ROC Curve (AUC), F1-Score, Recall, and Precision. These metrics provide a comprehensive evaluation of model performance, reflecting not only overall correctness (ACC), but also the model's ability to correctly identify positive cases (Recall), maintain precision in predictions (Precision), and balance between them (F1-Score). The AUC further measures the model's discrimination capability across all classification thresholds, which is particularly valuable in our class imbalance task. To assess statistical significance, we conduct AUC bootstrap tests. Additionally, we evaluate inference efficiency by measuring the average inference time (IT) per 10 cases, reported in seconds.

\subsubsection{Comparison Results}
As shown in Table~\ref{table1}, our proposed model achieves the best overall performance in subclassification while maintaining high computational efficiency. Among the comparison methods, Diff3Dformer delivers strong results with an average Accuracy, AUC, Recall, and F1-Score of 96.78\%, 97.25\%, 90.73\%, and 94.06\%, respectively. In contrast, our CECT-Mamba outperforms all competitors, achieving the highest Accuracy (97.41\%), AUC (98.60\%), Recall (93.13\%), and F1-Score (95.48\%), demonstrating superior diagnostic robustness. In terms of inference speed, our method achieves the second-fastest runtime with 0.85 seconds on average to process 10 cases, outperforming all LSTM-based and Transformer-based methods, and is only surpassed by ResNet. 
It is worth noting that the LSTM-based method outperforms other comparative approaches, highlighting the importance of temporal features in this subtype classification task. However, this improvement comes at the cost of significantly higher computational overhead. Similarly, Diff3Dformer incurs substantial computational expense during training due to its reliance on a reconstruction-based pretraining strategy. In contrast, our proposed method achieves the best overall performance by striking a favorable balance between computational efficiency and classification accuracy.
Moreover, a statistical significance bootstrap test shows that our method yields a significantly higher AUC than all comparison methods, with p-values less than 0.05. These results demonstrate that our approach effectively captures both spatial and temporal cues through global modeling, enabling accurate and efficient tumor diagnosis.

\subsection{Visualization}
We select seven comparative methods for the visual evaluation of our dataset. Grad-CAM~\cite{num32} is employed to visualize the regions of interest attended by the models. For consistency, we use the same slice from the CT scans of a patient across three imaging phases. We highlight a particularly challenging case in which the radiologist initially diagnosed the tumor as PDAC based on CECT images. However, postoperative pathological examination, the clinical gold standard, confirmed that the true diagnosis was a PNET. Among all comparison methods, only our method correctly predicts the tumor as a PNET, while all other methods misclassify it. Notably, as shown in Fig.~\ref{fig3}, our model focuses on tumor regions that are more compact and localized. This suggests that the tumor region highlighted by our model in the Grad-CAM visualization may correspond to diagnostically critical areas for identifying PNET, indicating potential clinical relevance and demonstrating superior performance compared to other state-of-the-art methods.

% \subsection{Ablation Study}
\subsection{Ablation Study}
To evaluate the effectiveness of our proposed sub-modules in CECT-Mamba, we compared five baseline networks. The first baseline (denoted as ``Basic'') is to remove all components from our model, retaining only four down-sampling blocks composed of convolutional layers. Incorporating the SCI module into ``Basic'' results in ``C1''. Then, we introduce the DHCM module with only spatial modeling ($M_S$) to ``C1'' yields ``C2'' and further incorporate temporal modeling ($M_T$) without the SimR module results in ``C3''. Adding the SimR module produces ``C4''. Finally, integrating the MGF module leads to our full CECT-Mamba method. 
% The results, presented in Table~\ref{table2}, demonstrate progressive performance improvement across all metrics with the sequential addition of each proposed module to ``Basic''.
The results in Table~\ref{table2} show a consistent performance improvement with the progressive integration of each component. For instance, adding the SCI module alone (C1) brings a large gain in AUC from 83.47\% to 92.37\% and improves Recall from 58.58\% to 78.30\%. With the introduction of spatial and temporal modeling in DHCM (C2 and C3), AUC further increases to 96.78\% and Recall reaches 90.60\%. Incorporating the SimR module (C4) pushes the Recall to 92.18\% and boosts overall classification robustness. Our full model achieves the highest performance across all metrics, with 97.41\% ACC, 98.60\% AUC, 93.13\% Recall, and 95.48\% F1-Score, confirming the complementary contributions of each sub-module.

\section{Conclusion}
In this work, we propose CECT-Mamba, a novel framework that leverages multi-phase contrast-enhanced CT for the classification of pancreatic tumor subtypes. Our framework excels at capturing fine-grained, inter-phase global enhancement dynamics, effectively addressing the challenges for subclassification posed by the high inter-class similarity and atypical enhancement patterns in different subtypes of pancreatic tumors.
Inspired by the clinical workflow of radiologists, our pipeline first localizes the tumor region, followed by the DHCM module, which incorporates Mamba to perform simultaneous spatial and temporal modeling. This design enables the extraction of both comprehensive volumetric representations and contrast-enhancement dynamics of the tumor. Additionally, the SCI and MGF modules are introduced to alleviate information loss and enhance the spatial and semantic complementarity of the 3D features. 
Extensive experiments on a dataset of 270 patients with multi-phase CECT scans demonstrate the effectiveness and superior performance of our method. Ablation studies further confirm the individual contribution and necessity of each proposed module.

\subsection*{Acknowledgement}
This work was supported by the National Natural Science Foundation of China (NSFC) under Grant 82330060.
{
    \small
    \bibliographystyle{ieeenat_fullname}
    \bibliography{main}
}

\end{document}